\documentclass[journal]{IEEEtran}
\usepackage{graphicx}
\usepackage{subfigure}
\usepackage{amsmath}
\usepackage{epstopdf}
\usepackage{booktabs}
\usepackage{algorithm}
\usepackage{algorithmic}
\usepackage{amsfonts,amssymb}
\usepackage{multirow, makecell}
\usepackage{amsfonts,amssymb}
\usepackage{url}

\usepackage{cite}
\usepackage{amsmath}
\usepackage{algorithmic}
\begin{document}
%
\title{AIDA: Legal Judgment Predictions for Non-Professional Fact Descriptions via Partial-and-Imbalanced Domain Adaptation}
%
%
%

\author{
	Guangyi~Xiao,~\IEEEmembership{Member,~IEEE,}
	Xinlong~Liu,~
	Hao~Chen,~
	Jingzhi~Guo,~\IEEEmembership{Member,~IEEE,}
	Zhiguo~Gong,~\IEEEmembership{Senior~Member,~IEEE}
	\thanks{G. Xiao, X. Liu, H. Chen are with the College of Computer Science and Electronic Engineering, Hunan University, Chansha, China 410082 E-mail:guangyi.xiao@gmail.com, xinlongliu@hnu.edu.cn, chenhao@hnu.edu.cn}
	\thanks{J. Guo, is with the Department of Computer and Information Science, University of Macau, Macau SAR 999078, China E-mail:jzguo@umac.mo}
	\thanks{Z. Gong, is with the State Key Laboratory of Internet of Things for Smart City, and the Department of Computer and Information Science, University of Macau, Macau SAR 999078, China E-mail:fstzgg@umac.mo}
	
}

%
%

\markboth{}%
{Shell \MakeLowercase{\textit{et al.}}: Bare Advanced Demo of IEEEtran.cls for IEEE Computer Society Journals}
%



\maketitle

\begin{abstract}
In this paper, we study the problem of legal domain adaptation problem from an imbalanced source domain to a partial target domain. The task aims to improve legal judgment predictions for non-professional fact descriptions. We formulate this task as a partial-and-imbalanced domain adaptation problem. Though deep domain adaptation has achieved cutting-edge performance in many unsupervised domain adaptation tasks. However, due to the negative transfer of samples in non-shared classes, it is hard for current domain adaptation model to solve the partial-and-imbalanced transfer problem. In this work, we explore large-scale non-shared but related classes data in the source domain with a hierarchy weighting adaptation to tackle this limitation. We propose to embed a novel p\textbf{A}rtial \textbf{I}mbalanced \textbf{D}omain \textbf{A}daptation technique (AIDA) in the deep learning model, which can jointly borrow sibling knowledge from non-shared classes to shared classes in the source domain and further transfer the shared classes knowledge from the source domain to the target domain. Experimental results show that our model outperforms the state-of-the-art algorithms.
\end{abstract}

\begin{IEEEkeywords}
Legal Intelligence, Partial Domain Adaptation, Imbalanced Domain Adaptation, Non-professional Fact Description, Fact Description, Legal Judgment Prediction.
\end{IEEEkeywords}

%
\IEEEpeerreviewmaketitle

\section{Introduction}
%
%
%
%
\IEEEPARstart{T}{he}
availability of large amounts of law-related data enables substantial progress of  information intelligence in the law domain, which generate several legal assistant systems to support users in finding relevant cases by giving some query\cite{Chen2013} or looking for some applicable law articles for a given case\cite{Liu2005}. Among all functions, legal judgment prediction is one of the most important ones in such a system, which is to predict approximate charges and applicable law articles by giving a fact description \cite{zhong2018legal}. However, in order to effectively use such a system, the user has to write the fact description in a professional format because the prediction algorithm of the system is often trained using a dataset of \textit{professional fact descriptions (PFD)}. This feature significantly restricts the common users from using the system because they can only give their cases as \textit{non-professional fact descriptions(NPFD)}. In this paper, we propose a novel domain adaptation technique to solve the problem.
\par
\begin{figure}[t]
	\begin{minipage}[b]{1.0\linewidth}
		\centering
		\includegraphics[width=3.2in,height=2.0in]{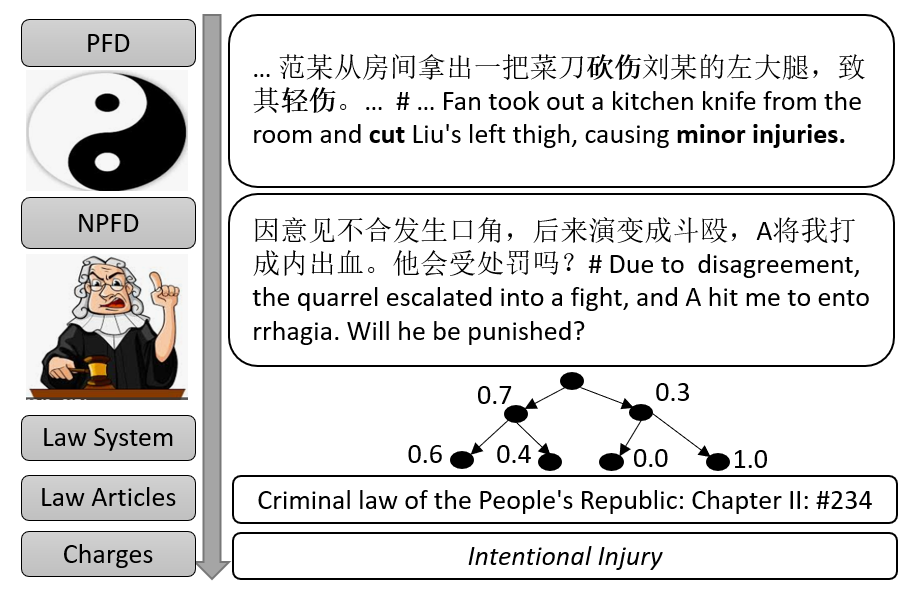}
		\caption{An illustration of the judicial logical of human judges in civil law system transferring from \textit{PFD} to \textit{NPFD}. }
		\label{Fig. 1}
	\end{minipage}
\end{figure}

\subsection{Motivations}
\textit{Non-professional fact description (NPFD)} is referred to the legal fact writing by non-professional people, whose expression of a fact may be much different from a professional description. Therefore, it is hard to build a reliable automatic prediction system for \textit{NPFD} since the lack of the training data (labeled data) for the non-professional fact descriptions. On the other hand, massive labeled \textit{PFD} data are released by different law-related organizations. For example, Chinese government published a large amount of court-related data ever since 2013 \footnote{https://wenshu.court.gov.cn/}, where each case is well-structured and divided in a professional manner into several parts such as fact and penalty. The \textit{PFD} dataset provides
a massive legal judgment resource for training a prediction algorithm, thus, is popularly used in several works, such as \cite{zhong2018legal}\cite{luo2017learning}\cite{hu2018few}, as training data for \textit{charge prediction tasks} and \textit{article prediction tasks}. To solve the problem of \textit{NPFD} predictions, one natural idea is to use the advanced unsupervised domain adaptation techniques\cite{long2017deep} to transfer knowledge from \textit{PFD} domain to \textit{NPFD} domain via sharing parameters of word embedding and text encoder. The key for the transfer learning technique to work successfully is the existence of latent invariant features crossing the domains. In our scenario, because the words or phrases used in \textit{PFD} and \textit{NPFD}  can be very different, we are going to exploit the word embedding technique to reduce the difficulty in looking for invariant features between the two domains. \par

Currently, \cite{Conneau2018Supervised} showed BiLSTM trained on the Stanford Natural Language Inference (SNLI) dataset yielded the state-of-the-art result compared with all existing alternatives of unsupervised world embedding techniques. In fact, word embedding techniques can only solve the domain mismatch by extending the sparse words used in the two domains into a common external context but could not learn the invariant features crossing them, which limits the improvement.
To solve the problem, in this work we propose to combine this advanced transferable text encoder\cite{Conneau2018Supervised} and high-level domain adaptation\cite{long2017deep} as a \textbf{strong baseline} for an unsupervised domain adaptation task. \par

Several techniques for domain adaptation are proposed in recent years.

\textbf{Sparse Partial Domain Shift.} As we mentioned, the fact descriptions for \textit{PFD} and \textit{NPFD} are much different. An example is as shown in Fig \ref{Fig. 1}, where both word \textit{“(cut)”} and \textit{“(minor injuries)”} are seldom used in \textit{NPFD}, it can be regarded as an indicator for \textit{intentional injury} in \textit{NPFD}. For \textit{NPFD}, people usually use spoken language to express the same semantics, which is also confirmed in \cite{luo2017learning}. \par

Furthermore, many practical applications of \textit{NPFD} only need to support predictions in a very  specific area (a subset of the broad law domain). Then the \textit{PFD} dataset can be divided to two parts: (1) \textit{Shared PFD} which shares the same label space of \textit{NPFD}, and \textit{Non-shared PFD} which shares no label space of \textit{NPFD}.

The previous domain adaptation techniques as proposed in \cite{long2017deep} often ignore the \textit{None-shared PFD},  such that the non-shared classes data in(\textit{PFD} domain) may bring much noise in extracting invariant features crossing the two domains, thus, lead to a lower performance in the prediction.
The only exception is partial domain adaptation (PDA) \cite{Cao_2018_CVPR}, which is the domain adaptation that the source label space subsumes the target label space for safe transfer of the unknown target space. However, PDA has several limitations: (1) bad performance for sparse shared-classes in source domain, (2) bad performance for large-scale non-shared classes in source domain.
\textbf{This paper argue the \textit{None-shared PFD} is helpful to improve traditional domain adaptation when some classes of \textit{Shared PFD} is sparse.}
\par

\textbf{Source Dataset Imbalance.} The distribution of various \textit{charges} and \textit{law articles} are quite imbalanced in \textit{PFD} dataset\cite{zhong2018legal}. In statics, the top 10 most frequent charges cover as much as 78.1\% cases while  the top least frequent 50 \textit{charges} only cover about 0.5\% cases.
This will cause a severe over-fitting problem to the categories with few cases. \par

\begin{figure*}[t]
	\begin{minipage}[b]{1.0\linewidth}
		\centering
		\includegraphics
		[width=7in]
		{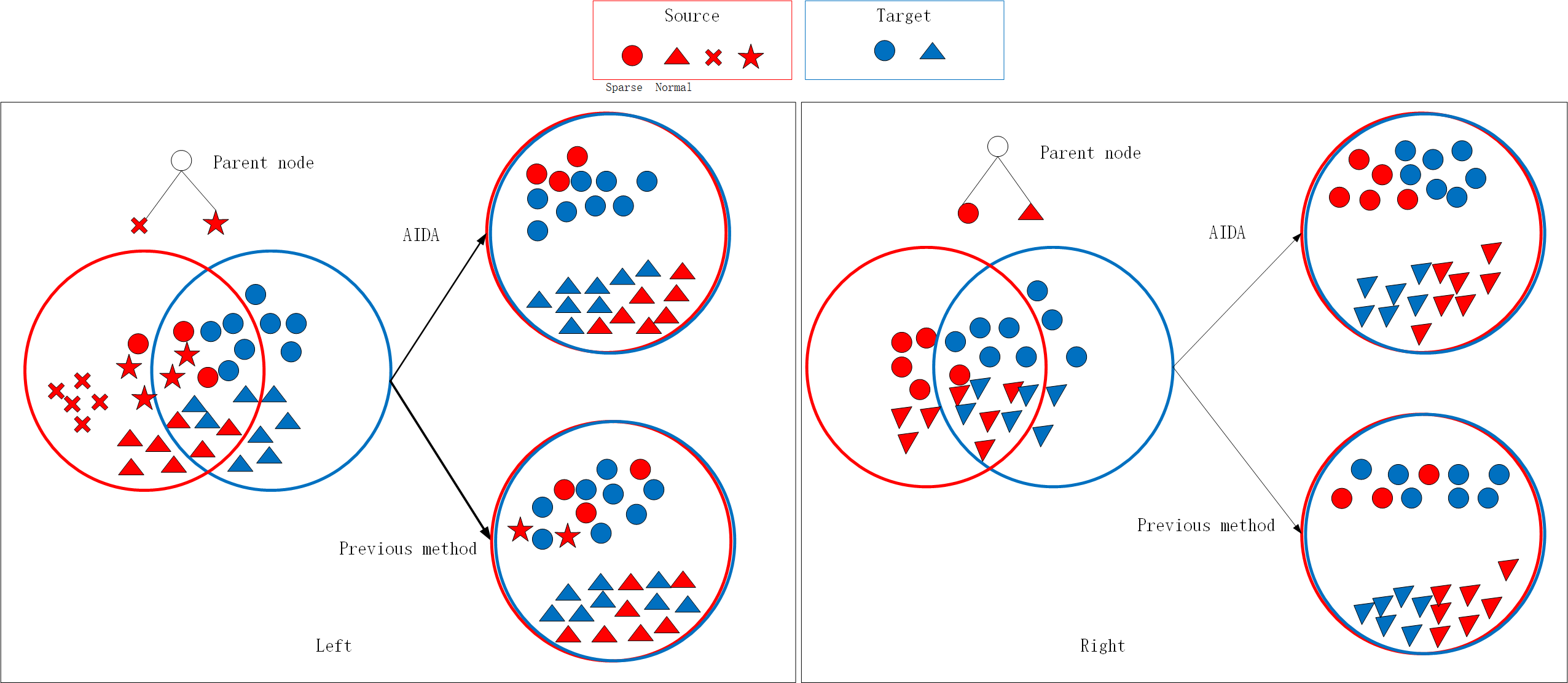}
		\caption{Partial-and-imbalanced domain adaptation (AIDA) is a generalized setting of domain adaptation where the source label subsumes the target label space and the source domain is imbalanced. (1) In the left of figure, we consider the sibling relationship between the red star and the red cross, belonging to the same parent node. The previous method ignores the sibling relationship between them, which may cause the red star to be mixed with the red circle, so that samples not belonging to the shared-class are misclassified to the target domain. In our solution, considering the sibling relationship between the red star and the red cross, the red cross can keep the red star away from the overlapping area. (2) In the right figure, we assume that the red circle represents the sparse shared-class in the source domain, and the red triangle represents the normal class. Without the help of the sibling' knowledge, due to the sparsity of the sample, the red circle does not have sufficient information to assist the target domain during the migration process, resulting in information loss. In our approach, we effectively use the samples of the large-scale sibling classes associated with them to help sparse shared-classes and enhance the robustness of the model.}
		\label{Fig. 2}
	\end{minipage}
\end{figure*}
\par

\subsection{Contributions}
Despite a large body of work in each of two areas, the \textbf{Partial-and-Imbalanced Domain Adaptation (PIDA)} problem of the link between \textbf{Sparse Partial Domain Shift} and \textbf{Source Data Imbalance} in unsupervised domain adaptation has not been fully explored. \par

Firstly, to solve the fundamental \textit{shared domain shift} problem, a basic idea is to confuse the \textit{Shared PFD} domain and \textit{NPFD} domain\cite{long2017deep} with adversarial domain adaptation. We extend the advanced Conditional Domain Adversarial Network (CDAN)\cite{long2018conditional} to \textit{shared domain adaptation} in a flexible way, which can support any flexible specific sub-area classifier (shared-classifier). So we propose the flexible Shared Adversarial Domain Adaptation (SADA) network with a Mask Filter Layer (MLL). \par

Secondly, to improve SADA by solving the advanced PIDA, we proposed a Hierarchy Weighting Adaptation (HWA) network which make use of the sibling relationship in the sparse shared-classes and large-scale non-shared-classes in \textit{PFD}. The hierarchies (sibling relationship) of \textit{charge} classes and \textit{applicable articles} classes can be easily extracted from the China Civil Law System \footnote{\url{http://www.npc.gov.cn/wxzl/gongbao/2000-12/17/content_5004680.htm}}.
Usually, the classes of \textit{charges} are dependent on the \textit{law articles}\cite{zhong2018legal}. For example, we know that \textit{“(organized robbery crime)”} is related to \textit{“(the robbery)”} and \textit{“(the snatch)”}, these three charges are attributed to \textit{the property infringement} in China Civil Law System. Compared with the other two categories, \textit{the crowd rob} is a sparse shared-class in \textit{Shared PFD}. \par

Knowing sibling relationship could allow us to borrow "knowledge" from relevant classes in \textit{Non-shared PFD} so that we can conger the over-fitting problem of sparse classes in \textit{Shared PFD}. Over-fit problem of sparse shared-classes is a challenge problem for adversarial domain adaptation since the over-distorting of feature extractor with sparse shared-classes in adversarial. Therefore, we propose to maintain the sibling connects between all classes in \textit{PFD} to reduce the over-distorting between two domains especially for sparse shared-classes. This paper presents a new Hierarchy Weighting Adaptation network (HWA) to address the technical difficulties of PIDA. This HWA has two novel characteristics to improve SADA: (1) maintained the shared-classes and non-shared-classes with semantic structure consistency in the feature representation space to reduce the over-distorting of two domains especially for sparse shared-classes; (2) the feature representation of non-shared classes in \textit{Non-shared PFD} is learned as required by those most sparse sibling classes in \textit{Shared PFD} with a novel weighting schema of sparseness.   \par

Our contributions in this paper can be summarized as follows:
\begin{itemize}
\item This is a first method to embed the transfer learning technique into the word embedding technique to solve the word shift problem of fact descriptions between \textit{PFD}  and \textit{NPFD} with a adversarial domain adaptation.

\item The PIDA problem, and the deep issues such as over-distorting of feature extractor in adversarial, negative transfer learning of unnecessary non-shared classes are ended with HWA network, which enables sparse classes to borrow knowledge from their sibling classes in large-scale non-shared classes by semantic structure consistency. A trade-off between the positive and negative effect of the representation learning of non-shared classes is designed.

\item Extensive experiments are conducted on two public real-world datasets. The results show the effectiveness and efficiency of our model for legal judgment predictions of \textit{NPFD}. Meanwhile, we have released the datasets and our source code\footnote{\url{https://law-intelligence.wixsite.com/domain-adaptation}}.
\end{itemize}
\par

\section{Related works}
Our research builds on previous works in the field of legal intelligence, unsupervised domain adaptation, and partial domain adaptation.
\subsection{Legal intelligence}
Legal intelligence has been studied for decades and most existing work formalizes this task under the text classification framework \cite{Kort1957Predicting}\cite{Segal1984Predicting}\cite{Lauderdale2012The}. Since machine learning has been proven successful in many areas, researchers begin to formalize Legal intelligence with machine learning methods. \cite{sulea2017exploring} develop an ensemble system to improve the related text classification performance.\par
Due to the publication of large criminal cases\footnote{https://wenshu.court.gov.cn/} and the successful application of deep neural networks in natural language processing \cite{kim2014convolutional}\cite{lai2015recurrent}, legal intelligence has become a hot topic in recent years.\par
Researchers began to improve the level of intelligence in the legal field by combining neural models with legal knowledge. For example, \cite{luo2017learning} propose an attention-based neural network method that jointly models charge predictions and relevant article extraction in a unified framework. Besides, \cite{ye2018interpretable} explore charge labels to solve the non-distinctions of fact descriptions and adopt a Seq2Seq model to generate court views and predicted charges. \cite{hu2018few} introduce discriminative legal attributes into consideration and propose a novel attribute-based multi-task learning model for \textit{charge predictions} of \textit{PFD}.  \cite{zhong2018legal} address multiple subtasks of judgement predications of \textit{PFD} with a topological learning framework.\par
Although these efforts have promoted milestones of legal intelligence, exist works ignores the non-professional people's demands for legal intelligence on \textit{NPFD}.
\subsection{Unsupervised domain adaptation}
The aim of domain adaptation \cite{pan2010survey} is to transfer knowledge between two domains by reducing domain shift. Deep networks are remarkable to learn feature representation of transferability \cite{yosinski2014transferable}. Currently, combining the theories of domain adaptation \cite{ben2010theory} to feature representation in deep network is the main research direction for domain adaptation.\par
Various algorithms have been proposed for domain adaptation in natural language process\cite{ding2018domain}\cite{li2018hierarchical}\cite{Zheng2017End}. Based on the techniques used in domain adaptation, these methods, in general, achieved in three different ways:(1)Discrepancy-based domain adaptation use Maximum Mean Discrepancy(MMD) criterion to reduces the shift between the two domains\cite{long2017deep}\cite{tzeng2014deep}\cite{long2015learning}; (2)Adversarial-based method refers to introduce adversarial technology inspired by the Generative adversarial nets(GAN) to find transferable representations that is applicable to both the source domain and the target domain\cite{ganin2014unsupervised}\cite{tzeng2017adversarial} or (3) Pixel-level domain adaptation method refers to translate the source images to target images \cite{liu2017unsupervised}\cite{sankaranarayanan2018generate}\cite{hoffman2018cycada}.\par
Recently, adversarial-based methods are remarkable based on works to improve the domain discriminator or the procedure of adversarial learning. ADDA design two feature representation for source domain and target domain respectively \cite{tzeng2017adversarial}. With the same feature representation for two domains, CDAN use the feature representation and the classier prediction to condition domain discriminator \cite{long2018conditional}. Applied the more effect of the divergence of with two hypothesis \cite{ben2010theory}, Saito et al. \cite{ding2018domain} proposed the adversarial domain adaptation with two classifiers. GPDA follows the procedure of the adversarial domain adaptation of two classifiers with a deep max-margin Gaussian processes \cite{kim2019unsupervised}.  \par
However, these methods are based on exacted match of label spaces for two domains or the data distribution of the source domain is balanced.

\begin{figure*}[t]
	\begin{minipage}[b]{1.0\linewidth}
		\centering
		\includegraphics
		[width=6in]
		{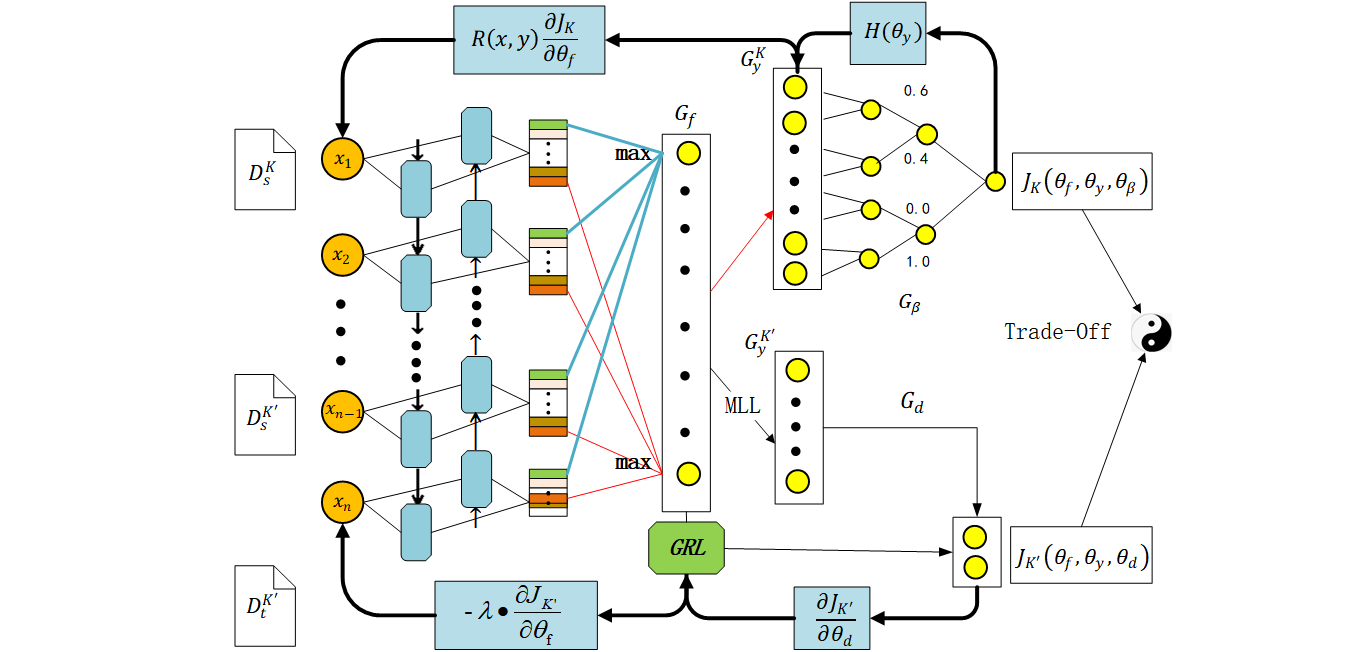}
		\caption{The architecture of AIDA consists two networks and a trade-off algorithm. The bottom network is flexible shared domain adaptation network (SDAN) for $D_s^{k'}$ and $D_t^{k'}$, and the upper network is hierarchy policy network (HPN) for $D_s^k$, where $D_s^k$ and $D_s^{k'}$ are labelled all classes and shared classes samples of \textit{PFD} respectively and $D_t^{k'}$ are unlabelled target classes of \textit{NPFD}. The AIDA algorithm tries trade-off two networks with two circle systems using the representation learning of the \textbf{required} classes and \textbf{important} examples of \textit{Non-shared PFD}.}
		\label{Fig. 3}
	\end{minipage}
\end{figure*}

\subsection{Partial domain adaptation}
While the standard domain adaptation evolves prosperously, it still needs to assume two domains shared the same label space. This assumption does not hold in partial domain adaptation (PDA), which transfers models from large-scale-classes to special domain and assume the label space of target domain is unknown. There are four main efforts to this problem. Selective Adversarial Network (SAN)\cite{Cao_2018_CVPR} adopts multiple adversarial networks with weighting mechanism to select out source examples in the non-share classes. Partial Adversarial Domain Adaptation (PADA) \cite{Cao2018} improves SAN by employing only one adversarial network and further adding the class-level weight to the source classifier. Importance Weighted Adversarial Nets (IWAN) \cite{Zhang_2018_CVPR} uses the sigmoid output of an auxiliary domain classifier without adversarial training to generate the probability of a source example. Example Transfer Network (ETN) \cite{Cao_2019_CVPR} use a progressive weighting scheme to quantify the transferability of source examples. These fundamental methods achieve milestones in partial domain adaptation tasks. \par

However, for partial-and-imbalanced domain adaptation, this methods may failed for sparse shared-classes since the negative transfer learning still exist for situation of the large-scale of non-shared classes, this will be justified in experiments.

\section{AIDA: pArtial-and-Imbalanced Domain Adaptation Network}
\subsection{Problem Definition}
In this paper, we propose a robust partial-and-imbalanced domain adaptation for \textit{NPFD} predictions. In the design, we assume that the \textit{PFD} label space $C^K_s$ is a superset of the \textit{NPFD} label space $C^{K'}_t$. We denote label space $C^{K'}_s(\subset C^K_s)$ in \textit{PFD} domain as the shared-classes with the target domain, such that $C^{K'}_s=C^{K'}_t$ but with different training datasets (\textit{PFD} and \textit{NPFD} respectively).
We further have a labelled dataset $D^K_s=\{(x^s_i,y^s_i)\}^{n_s}_{i=1}$ in \textit{PFD}, where $n_s$ denotes the size of $D^K_s$. Correspondingly, $D^{K'}_s=\{(x^s_i,y^s_i)\}^{n^\prime_s}_{i=1}$ denotes the shared-classes dataset in \textit{PFD} with size $n^\prime_s$, such that $D^{K'}_s\subset D^K_s$. In \textit{NPFD} domain, there is an unlabelled dataset $D^{K'}_t=\{x^t_i\}^{n_t}_{i=1}$ for the target classes $C^{K'}_t$ with size $n_t$.\par

The traditional transfer learning techniques assume that $C_s=C_t$ and source \textit{PFD} and target \textit{NPFD} are sampled from distributions $P$ and $Q$ respectively,  and $P \ne Q$. In our scenario (\textbf{partial-and-imbalanced transfer learning}), we have $P^{K'}_s \ne Q^{K'}_t$, where $P^{K'}_s$ denotes the distribution of the source labelled \textit{PFD} datasets w.r.t. shared-classes $C^{K'}_s$, $Q^{K'}_t$ denotes the distribution of the target unlabelled \textit{NPFD} dataset w.r.t. label space $C^{K'}_t$. Then, we consider the following real situation: (1) There is a mismatch between $P_s^{K'}$ and $Q_t^{K'}$; (2) $D^K_s$ is imbalanced and $D^{K'}_s$ contains sparse classes. In fact, the second problem is often ignored in the existing transfer learning techniques, which we will tackle significantly in this paper.\par

The existing \textbf{Shared Domain Adaptation (SDA)} techniques implement transfer ${D^{K'}_s}\rightarrow {D^{K'}_t}$ by  learning invariant features between ${D^{K'}_s}$ and ${D^{K'}_t}$ in the shared label space $C^{K'}$. \textbf{Partial Domain Adaptation (PDA)} techniques assume the target label space $C^{K'}$ is unknown, and incorporate the model to learn the weight (possibility) of samples of  ${D^{K}_s}$ located in the shared label space ${D^{K'}_s}$. In this paper, the target label space $C_t^{K'}$($\subset C^{K}_s$) is known to us but with sever imbalanced classes. As we will show in the experimental section, neither of existing techniques could handle the imbalance problem of the source training data, though it is very practical in the real world. As the result, the sparse classes are easily ignored by the algorithms. \par

In our proposed solution, we want to build a robust deep learning solution to transferring knowledge from $D^K_s$ to $D^{K'}_t$ via middle dataset $D^{K'}_s$, $D^K_s\rightarrow{D^{K'}_s}\rightarrow{D^{K'}_t}$ named \textbf{pArtial-and-Imbalanced Domain Adaptation (AIDA)}, ${D^{K'}_s}\rightarrow{D^{K'}_t}$ (the direct transfer from the sheared classes of the source domain) as \textbf{Shared Domain Adaptation (SDA)}, and ${D^{K}_s}\rightarrow{D^{K'}_t}$ (the transfer from the overall source domain to the target domain) as \textbf{Partial Domain Adaptation (PDA)}. The objective is not only to overcome the domain shift but also fight the imbalance of the training data.\par

\subsection{AIDA Framework}
Firstly, we introduce how to solve the fundamental problem of shared domain adaptation from the flexible \textit{Shared PFD} to target \textit{NPFD} in adversarial. Then we introduce how to improve shared adversarial domain adaptation by fixing advanced PIDA problem with large-scale \textit{Non-shared PFD}.\par

\subsubsection{Shared Domain Adaptation Network}
The fundamental problem of domain adaptation is to reduce the distribution shift between shared-classes of $P_s^{K'}$ and $Q_t^{K'}$. \par

Domain adversarial networks\cite{ganin2016domain} tackle this problem by learning transferable features in a two-player minimax game: the first player is a domain discriminator $G_d$ trained to distinguish the feature representations of the  \textit{Shared PDF} from the \textit{NPDF}, and the second player is a feature extractor (generator) $G_f$ trained simultaneously to fight the domain discriminator. The domain-invariant features \textbf{f} are learned in a minimax optimization procedure: the parameters $\theta_f$ of the feature generator $G_f$ are trained by maximizing the loss of domain discriminator $G_d$, while the parameters $\theta_d$ of the domain discriminator $G_d$ are trained by minimizing the loss of the domain discriminator $G_d$. Note that the goal is to learn a source classifier that transfers to the target, hence the loss of the source classifier $G_y$ is also minimized. \par
Conditional domain adversarial network (CDAN) \cite{long2018conditional} exploits the classifier predictions $G_y$ to enhance domain adversarial adaptation using such rich discriminative information. It conditions domain discriminator $G_d$ on the classifier prediction \textbf{g} through the multi-linear map:
\begin{align}
T_\otimes (\textbf{h})= \textbf{f} \otimes \textbf{g} \label{(1)}
\end{align}
where $\textbf{h}=(\textbf{f},\textbf{g})$. This leads to the optimization problem proposed in CDAN \cite{long2018conditional}:
\begin{align}
\mathcal{J}(\theta_f, \theta_y, \theta_d)= \mathbb{E}_{(x_i,y_i)\sim D_s} {L_y(G_y(x_i),y_i)} \nonumber \\
-\lambda \cdot \mathbb{E}_{x_i\sim D_a}
{L_d(G_d(T_\otimes (\textbf{h})),d_i)} \label{(2)}
\end{align}
where $D_a=D_s\sqcup D_t$ is the union of the source and target domains and $n_a=|D_a|$, $d_i$ is the domain label, $L_y$ and $L_d$ are the cross-entropy loss functions. Note that $\textbf{h}=(\textbf{f},\textbf{g})=(G_f(x_i), G_y(G_f(x_i))$ is the joint variable of domain-specific feature representation $\textbf{f}$ and classifier prediction $\textbf{g}$, which is significant adversarial domain adaptation as shown in \cite{long2018conditional}.\par

\textbf{Mask Filter Layer (MLL).} Specifically, for the target \textit{NPFD}, we only need the classification task for $K'$ target classes which is a sub-set $K$-classes. In flexible, the parameters of the flexible $K'$-classes classifier are shared with all $K$-classes classifier by a \textbf{Mask Filter Layer (MLL)}. Since the flexible $K'$-classes of shared domain adaptation (SDA), ${D^{K'}_s}\rightarrow{D^{K'}_t}$, from the \textit{Shared PFD} to specify \textit{NPFD} is required in real practice. \par

Suppose $G_K(G_f(x))\in R^K$ is the vector score of the last fully-connect layer, $G_y^K(G_K(G_f(x)))$ is the $K$-classes predict probability with softmax of $G_K(G_f(x))$. In order to focus on specified $K'$ subset label space, we need a mechanism to filter non-shared-classes in the classifier.
We use a \emph{MLL}:
\begin{align}
z^{K'}_j =
\begin{cases}
z^K_j& M_j = 1\\
-\infty& M_j = 0
\end{cases}
,1\leq j\leq K \label{(3)}
\end{align}
where $z^K \in {R}^K$ is output of vector score of $G_K(G_f(x))$, and the mask $M\in {R}^K$ indicates one hot representation of the shared-classes in the source domain. The value of vector $z_j$ corresponding to non-shared-classes are set to $-\infty$. After that, $z$ is directed forwarded to softmax layer. Therefore the origin classifier prediction $G_y^K(x)$ of all $K$-classes is extended to flexible shared classifier prediction $G_y^{K'}(x)$ of $K'$-classes with $MLL$:
\begin{align}
G_y^{K'}(x) = softmax(MLL(G_K(G_f(x)))) \label{(4)}
\end{align}
where all of $G_y^{K'}(x)$, $G_y^K(x)$ and $G_K(x)$ shared the same parameters $\theta_y$.
 \par
\textbf{Shared Domain Adaptation Network (SDAN).}  SDAN extends CDAN in a flexible way to allow the target  label space to be a subset of the source label space $C_t^{K'}\subset C_s^K$. Thus, large mount of training data from the source domain can be out of the class space of the target domain, and the source \textit{PFD} is divided to \textit{Shared PFD} and \textit{Non-shared PFD}, So we propose the SDAN with \textit{Shared PFD} and \textit{NPFD} as:
\begin{align}
\mathcal{J}_{K'}(\theta_f, \theta_y, \theta_d)= \mathcal{J}_{G_y^{K'}} - \lambda \cdot \mathcal{J}_{G_d^{K'}} \label{(5)}
\end{align}
where
\begin{align}
\mathcal{J}_{G_y^{K'}} = \mathbb{E}_{(x_i, y_i)\sim D_s^{K'}}  {L_y(G_y^{K'}(x_i),y_i)},  \label{(6)} \\
\mathcal{J}_{G_d^{K'}} = - \mathbb{E}_{x_i\sim D_s^{K'}} {log(G_d^{K'}(T_\otimes (\textbf{h})))} \nonumber \\
- \mathbb{E}_{x_i\sim D_t^{K'}} {log(1-G_d^{K'}(T_\otimes (\textbf{h})))}, \label{(7)}
\end{align}
$\textbf{h} = (\textbf{f},\textbf{g})=(G_f(x_i), G_y^{K'}(x_i))$, and $L_y$ is the cross-entropy loss function.\par
\subsubsection{Hierarchy Policy Network}
The key PIDA problem is the sparse \textit{Shared PFD}, this will lead to severe over-fitting problem for SDAN. Our idea is to fix over-fitting problem of sparse shared-classes by exploiting large-scale non-shared classes of $PFD$ with sibling relationship, then to improve SDAN. \par

Over-fitting problem of sparse \textit{Shared PFD} is a challenge problem for adversarial domain adaptation, since the \textit{adaptivity} between two domains for sparse shared-classes may be huge since the over-distorting of feature generator $G_f$. Fig. 2 shows the intuition illustration for the sparse over-fitting problem. We propose to maintain the semantic hierarchy consistency between \textit{Shared PFD} and large-scale \textit{Non-shared PFD} for the feature extractor $G_f$ to reduce the over-distorting for \textit{Shared PFD}. Therefore we integrate \textit{Shared PFD} and large-scale non-shared \textit{Non-shared PFD} as the \textbf{all $K$-classes \textit{PFD}}, then a hierarchy policy network to mitigate it:
\begin{align}
\mathcal{J_K}(\theta_f, \theta_y)=\mathcal{J}_{G_y^K} +  H(\theta_y) \label{(8)}
\end{align}
where the first item is the cross-entropy loss of all $K$-classes classifier  for \textit{PFD}, and the second term is the hierarchy constraint between the sibling connects of all \textit{PFD} to mitigate the over-distorting of $G_f$.
\par
To solve PIDA in SADA, our goal is to combine equation (5) and equation (8) together in a integrated framework. \par

\textbf{Sibling Relationship.} Therefore we constrain the rich classes and sparse classes with the semantic hierarchy consistency in feature representation space, that is, we should cluster the rich classes and sparse shared-classes together in feature representation space, if these two kinds of classes have the same parent abstract concept. Additionally, the \textit{stability} of the integrated weighing schema is important for optimization in such a imbalanced situation. Therefore, \textbf{hierarchy policy network (HPN)} exploits the rich data from the non-shared classes to promote those sparse shared-classes through their sibling relationships and hierarchy policy, thus to improve SDAN.
\par

To formulate the sibling relationships, we extract a three-level hierarchy for the integrated \textit{PFD}($D^K_s$) by using the civil law system. Suppose we have integrated $K$ leaf nodes corresponding to $K$ classes for \textit{charge prediction} task or \textit{law article prediction} task, which are grouped into $G$ super classes where $K\gg G$. Each leaf category \emph{k} is associated with a parameter vector $\theta_y^k\in \mathbb{R}^D$,
and each super-class node $j$ is associated with a vector $\theta_\beta^{j}\in \mathbb{R}^D$ where $1 \leq j \leq |G|$. We define the following generative model for $\theta_k$
\begin{align}
\theta_\beta^{j} \sim \mathcal{N}(0, \delta\cdot I_D), \quad \theta_y^k \sim \mathcal{N}(\theta_\beta^{parent(k)},  \delta\cdot I_D) \label{(9)}
\end{align}

\vspace{-0.4em}
\noindent where $\mathcal{N}(\cdot)$ is the Gaussian distribution with diagonal covariance and $\delta$ is a global constant.
\par
\textbf{HPN.} Unlike priori tree network (PTN)\cite{NIPS2013_5029} treats each non-shared examples equality, our HPN algorithm try to achieve the objective with different weights of sparseness reward to learn police model $(\theta_f, \theta_y, \theta_\beta)$:
\begin{align}
\mathcal{J}_K(\theta_f, \theta_y, \theta_\beta)= \\
\lambda_2 \cdot\mathbb{E}_{(x_i,y_i)\sim D_s^K} R(x_i, y_i) \cdot L_y(G_y^K(x_i),y_i)  \nonumber \\ + \lambda_3 \cdot H(\theta_y) \label{(10)}
\end{align}
where the first item is the cross-entropy loss function over the rewarding $R(x_i, y_i)$ of sparse weighting sample $(x_i, y_i)$. We constrain the consistency hierarchy structure of all classes with the $L2$ distance to their parent parameters $\theta_\beta^{parent(k)}$:
\begin{align}
H(\theta_y) =  \frac{\sum_{k=1}^{K} ||\theta_y^k - \theta_\beta^{parent(k)||}}{2}  \label{(11)}
\end{align}
where $\theta_y^k$ is the parameters of the last fully-connect classifier layer in $\theta_y$. With this consistency tree hierarchy constrains, the sparse over-fitting problem can be mitigated.
 Let $C_j=\{k|parent(k)=j\}$, then
\begin{align}
\hat{\theta}_\beta^j = \frac{1}{\mid C_j \mid} \sum_{k\in C_j} \theta_y^k \label{(12)}
\end{align}

\vspace{-0.4em}
Therefore, for HPN,the loss function in Eq.(10) can be optimized by iteratively performing the following two steps: (1) minimizing $\mathcal{J}_K(\theta_f, \theta_y)$ over $\theta_f$ and $\theta_y$ keeping $\theta_\beta$ fixed with policy gradient descent; (2) estimating $\theta_\beta$ that average the quantity with Eq. (12) keeping $\theta_y$ fixed. \par

However, they will yield negative transfer learning for the feature representation learning of large-scale \textit{Non-shared PFD}. Therefore, we need to devise a new mechanism with two steps: (1) a new \textbf{sparse reward mechanism} for each class of \textit{Non-shared PFD} to measure the quantity of requirement of the sparse \textit{Shared PFD}; and (2) a \textbf{trade-off algorithm} to balance the benefit of semantic hierarchy consistency and the negative transfer of the feature learning of large-scale \textit{Non-shared PFD}.\par

\begin{figure}[t]
	\begin{minipage}[b]{1.0\linewidth}
		\centering
		\includegraphics[width=3.2in]{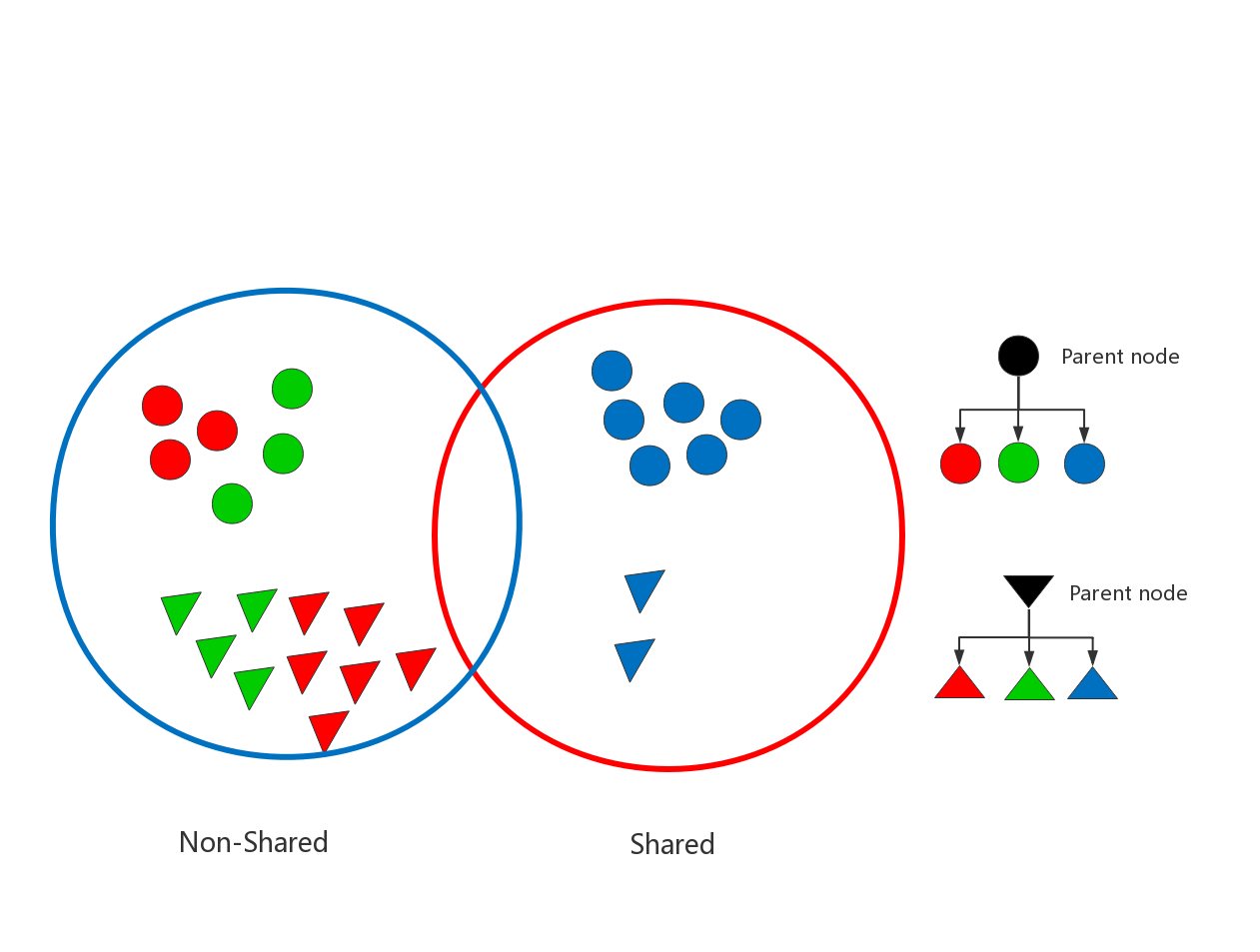}
		\caption{An illustration of sparse reward of non-shared classes in \textit{Non-shared PFD}. The representation learning of non-shared classes has negative transfer learning effect for shared domain adaptation, then what non-shared classes should to learn? The more sparseness of the dependent sibling requires more important weight to learning. The triangle classes in \textit{Non-shared PFD} need more important weight to learn than the cycle classes, since the triangle classes' dependent sibling classes are more sparse than the cycle classes.}
		\label{Fig. 4}
	\end{minipage}
\end{figure}

\textbf{Reward Shaping.}
One key contribution is a new \textbf{sparse reward mechanism} for each example of \textit{Non-shared PFD} to measure the quantity of requirement of the sparse \textit{Shared PFD}. Fig. 4 shows the intuition of sparse reward mechanism.  The representation learning of non-shared classes has negative transfer learning effect for shared domain adaptation, then the key question is what non-shared classes should to learn. The more sparseness of the dependent sibling requires more important weight to learning. The triangle classes in \textit{Non-shared PFD} need more important weight to learn than the cycle classes, since the triangle classes' dependent sibling classes are more sparse than the cycle classes.\par

\textbf{Class Reward.}
The sparse reward $r_1(y_i)$ denotes the weighting of each class $y_i$ in the all \textit{PFD}.
Suppose $S(y_i)$ is a set of sibling shared classes of $y_i$ in \textit{Shared PFD}. Then the sparse size of most sparse class $s(y_i)$ of the sibling shared class $S(y_i)$ is defined as:
\begin{align}
s(y_i)=
\begin{cases}
\infty& S(y_i) = None\\
\min \limits_{y\in S(y_i)} {N_y}& else
\end{cases}
\label{(13)}
\end{align}
where $N_y$ is the number of examples of each class $y$ in all \textit{PFD}. Smaller is better since sparse requirement.\par

We can calculate the class reward of sparse weighting $r_1(y_i)$ with a mini-batch normalization directly. We normalize the the class reward sparse weighting $r_1(y_i)$ for mini-batches of all dataset \textit{PFD}, thus the mini-batches of sparse weighting $r_1(y_i)$ is normalized with
\begin{align}
r_1(y_i)= \frac{\exp ^{-s(y_i)}}{\sum_{y \in B\sim D_s^K}{\exp ^{-s(y)}}} \label{(14)}
\end{align}
respectively, such that $\sum_{y_i}^{B\sim D_s^K}w(y_i)=1$ for approximating the class weighting of each example in \textit{PFD}.
\par

\textbf{Example Reward.}
The sparse reward $r_2(x_i)$ denotes the important weighting of each example $x_i$ in the all \textit{PFD}. Our example reward with joint variables of feature representation $\textbf{f}$ and noiseless shared-classifier $\textbf{g}$ is defined as:
\begin{align}
r_2(x_i)=1-G_d^{K'}(\textbf{f}, \textbf{g}) \label{15}
\end{align}
where $G_d^{K'}(\textbf{f}, \textbf{g})$ is the shared $K'$-classes domain discriminator trained to identify shared \textit{PFD} and target \textit{NPFD}.

We normalize the example reward $r_2(x_i)$ for mini-batches of two networks respectively, thus the mini-batches of example reward $R_2(X)$ are normalized with
\begin{align}
R_{2'}(X)=softmax_{X\sim D_s^{K}}(R_2(X)) \label{(16)}
\end{align}
respectively, such that $\sum_{x_i}^{X\sim D_s^{K}}r_{2'}(x_i)=1$ for approximating the joint dense distribution of \textit{PFD} domain and \textit{NPFD} domain receptively.
\par

\textbf{Final Reward.}
The final reward for each example of \textit{PFD} is defined as the product of class reward and example reward. The rewards are normalized globally across all examples in \textit{PFD}. The final reward is:
\begin{align}
R(x, y)= \alpha \times r_1(y) \times r_2(x) \label{(17)}
\end{align}
where $\alpha$ is the global normalization constant. When combined, benefits. \par

\begin{algorithm}[tb]
	\caption{\label{alg1}Trade-off Algorithm}
	\begin{algorithmic}[1] 
	    \WHILE{$cur\_iter < iternum$}		
		\STATE //{ setp 1: \textbf{SDAN}}.
		\STATE\textbf{Input}:{$D^{K'}_s=\{x^s,y^s\}$,$D^{K'}_t=\{x^t\}$}
		\STATE optimize $\theta_f, \theta_y,\theta_d$ by mini-batch of $D^{K'}_s$ and $D^{K'}_t$ with  Eq. \eqref{(5)} using SGR layer \;
		
		\STATE //{ setp 2: \textbf{HPN}}.
		\STATE\textbf{Input}:{$D^K_s=\{x^s,y^s\}$}
		\STATE calculate the sparse reward $R(x^s, y^s)$ with Eq. \eqref{(16)} \;
		\STATE optimize $\theta_f, \theta_y$ by mini-batch of $D^{K}_s$ with  Eq. \eqref{(10)} \;
		\STATE estimate  $\theta_\beta$ with Eq. \eqref{(12)}
		\ENDWHILE
		\STATE \textbf{return} {$\theta_f$, $\theta_y$, $\theta_d$, $\theta_\beta$}
	\end{algorithmic}
\end{algorithm}

\subsubsection{Trade-off Algorithm}
Another key contribution is to trade-off between the generalization power of the feature representation learning of non-shared classes and their negative transfer effect, and the detail training of \textit{AIDA} is presented in Algorithm 1. Overall, our algorithm tries to alternatively minimize the joint loss functions \eqref{(5)} and \eqref{(10)} iteratively in two cycle systems with the hierarchy policy network and shared domain adaptation network. The overall objective of \textit{AIDA} is to jointly learn \textit{SDAN} and \textit{HPN} by the objective:
The AIDA framework can be trained end-to-end by a minimax optimization procedure as follows, yielding a equilibria point solution $(\hat{\theta}_f, \hat{\theta}_y, \hat{\theta}_d, \hat{\theta}_\beta)$:
\begin{align}
\begin{matrix}
(\hat{\theta}_f, \hat{\theta}_y, \hat{\theta}_\beta) = \\
 \mathop{\arg\min}\limits_{\theta_f, \theta_y}  \mathcal{J}_{G_y^{K'}} - \lambda \cdot \mathcal{J}_{G_d^{K'}} + \lambda_2 \cdot \mathcal{J}_{G_y^K} + \lambda_3 \cdot H(\theta_y) \\
(\hat{\theta}_d) = \mathop{\arg\min}\limits_{\theta_d}  \mathcal{J}_{G_d^{K'}}
\end{matrix}
\label{(18)}
\end{align}
where first two items are the error of SDAN  and the last two items are the error of HPN. This is a minimax game to find a equalise point of $(\theta_f, \theta_y)$, where hyper-parameters $(\lambda_2, \lambda_3)$ are designed to balance the learning effect of \textit{SDAN} and that of \textit{HPN}. \par

Furthermore, the representation learning $\nabla_{\theta_f}\mathcal{J}_{G_y^K}(\theta_f, \theta_y)$ of non-shared classes has positive learning effect for HPN($\mathcal{J}_K(\theta_f, \theta_y)$), but has negative effect for the SDAN ($\mathcal{J}_{K'}(\theta_f, \theta_y, \theta_d)$). Although we mitigate it though policy gradient backward, to further solve it in trade-off, in contrast to set big weight $(1+\lambda_2)$ for representation learning $\nabla_{\theta_f}\mathcal{J}_{G_y^{K'}}(\theta_f, \theta_y)$ of the shared-classes $D_s^{K'}$ in \textit{Shared PFD}, we want to set a relative-small trade-off weight $\lambda_2$ for representation learning $\nabla_{\theta_f}\mathcal{J}_{G_y^K}(\theta_f, \theta_y)$ of the non-shared classes. In a word, AIDA algorithm tries to trade-off between: (1) the sibling power by leverage the examples of the non-shared but related classes in \textit{Non-shared PFD} domain, and (2) the potential negative transfer effect to the representation learning of non-shared classes in shared domain adaptation. \par

\subsubsection{Theoretical understanding}
Ben-David et al. \cite{ben2010theory} proposed the fundamental domain adaptation theory that bounds the expected error $\mathcal{J}_\mathcal{T}(h)$ of a hypothesis $h$ on the target domain by using three components: (1) expected error of $h$ on the source domain, $\mathcal{J}_\mathcal{S}(h)$; (2) $\mathcal{H}\nabla \mathcal{H}$-divergence $d_{\mathcal{H}\nabla \mathcal{H}}(\mathcal{S},\mathcal{T})$, measuring domain shift as the disagreement of two hypotheses $h, h' \in \mathcal{H}\nabla \mathcal{H}$; and (3) the \textit{adaptability} error $\lambda^*$ of the ideal joint hypothesis $h^*$ on both domains. The learning bound is
\begin{align}
\mathcal{J}_\mathcal{T}(h)\leq \mathcal{J}_\mathcal{S}(h) + \frac{1}{2}d_{\mathcal{H}\nabla \mathcal{H}}(\mathcal{S},\mathcal{T})+\lambda^*
\end{align}
where error $\lambda^*$ of the ideal joint hypothesis $h^*=min_h \mathcal{J}_\mathcal{S}(h)+\mathcal{J}_\mathcal{T}(h)$ is defined as
\begin{align}
\lambda^*=\mathcal{J}_\mathcal{S}(h^*)+\mathcal{J}_\mathcal{T}(h^*)
\end{align}
Noteworthily, most adversarial domain adaptation methods treated the error $\lambda^*$ as a constant, which align domain shift by the first two terms of errors \cite{liu2019transferable}. AIDA approach constraint the feature representation $G_f$ of sparse shared-classes with consistency connect with non-shared but related classes by hierarchy  weighting adaptation to keep $\lambda^*$ small, yielding lower bound for the target error. This will be justified in extensive experiments.

\subsubsection{Transferable Text Feature Extractor}
Bi-directional LSTMs(BiLSTM) with max pooling are employed as the text encoder due to its advanced ability in extracting semantic meanings on representative transferability\cite{Conneau2018Supervised}.\par
Given a word sequence $x=[X_1,X_2,...,X_T]$ of \textit{PFD} or \textit{NPFD} where $X_t$ is the input embedding of element t, with word embedding lookup and BiLSTM encoder $h=BiLSTM(x)$, the state of BiLSTM at position t is:
\begin{align}
h_t=[\overrightarrow{h_{t}},\overleftarrow{h_{t}}] \label{(21)}
\end{align}
where $\overrightarrow{h_t}$ and $\overleftarrow{h_t}$ are the state of the forward and backward LSTM at position t. We pre-trained the word embedding in \textit{PFD} dataset. After processing all time steps, we get a hidden state sequence $h=[h_1,h_2,...,h_T]$. Then we feed it into a max-pooling layer and select the maximum value over each dimension of the hidden units to form a size-fixed vector $v=[v_1,v2,...,v_d]$ as
\begin{align}
v_i=max(h_{1,i},h_{2,i},...,h_{T,i}), {\forall}i\in[1,d] \label{(22)}
\end{align}
where $d$ is the dimension of the hidden states. Therefore, the transferable text feature generator is defined as:
\begin{align}
G_f(x)=max(BiLSTM(x)) \label{(23)}
\end{align}

\par

\begin{table}
	\renewcommand{\arraystretch}{1.3}
	\caption{The statistic of different domain datasets.}
	\label{Tab.1}
	\centering
	\begin{tabular}{c|cc|cc}
		\hline
		Datasets &  \multicolumn{2}{c|}{$Source$}&\multicolumn{2}{c} {$Target$} \\
		\hline
		Task& Articles&Charges&Articles&Charges\\
		\hline
		Size & 200k & 200k& 5.3k & 5.3k \\
		\hline
		\#classes & 75&83& 20&20\\
		
	\end{tabular}
\end{table}

\begin{table}
	\renewcommand{\arraystretch}{1.3}
	\caption{The statistic of WOS-46985 and Quora datasets.}
	\label{Tab.2}
	\centering
	\begin{tabular}{c|c|c}
		\hline
		Datasets &  {$WOS-46985$}& {$Quora$} \\
		
		\hline
		Size & 4.68k & 1.37k \\
		\hline
		\#classes & 134&23\\
		
	\end{tabular}
\end{table}

\begin{table*}
	\subfigure[Acc. comparison results]{
		\begin{minipage}{1.0\linewidth}	
			\center
			\begin{tabular}{c|ccccc|ccccc}
				\hline
				Task &  \multicolumn{5}{c|}{$Article\quad Prediction$}   &\multicolumn{5}{c}{$Charge\quad Prediction$} \\
				\hline
				number & 50 & 100 & 200 & 500 & 1000 & 50 & 100 & 200 & 500 & 1000 \\
				\hline
				BiLSTM & $67.15$ & $72.30$ & $76.90$ & $79.25$ & $81.05$&$70.05$ & $72.15$ & $77.65$ & $80.35$&$80.15$\\
				TOPJUDGE & $78.05$ & $81.15$ & $81.65$ & $83.20$ & $82.90$&$79.75$ & $80.50$ & $81.65$ & $82.75$&$82.25$\\
				DDC & $66.90$ & $73.10$ & $76.00$ & $78.45$ & $77.20$ & $69.65$ & $73.45$ & $74.65$ & $79.00$ & $78.65$ \\
				DAN & $67.35$ & $74.95$ & $75.60$ & $82.70$ & $83.40$ & $71.70$ & $78.85$ & $81.50 $ & $82.25$ &$83.00$\\
				DANN& $70.50$ & $76.35$ & $74.5$ & $81.75$ & $82.40$ & $71.15$ & $74.65$ & $76.65$ & $81.05$&$81.45$\\
				JAN & $74.50$ & $78.40$ & $78.35$ & $83.05$ &$84.05$& $76.70$&$ 80.65$&$ 82.75$&$ 84.45$&$85.25$\\
				CDAN & $90.60$ & $91.85$ & $87.65$ & $89.60$ &$87.85$& $90.55$&$91.70$&$ 89.25$&$ 88.70$&$87.15$\\
				\hline
				PADA&$-$&$11.70$&$26.20$&$47.30$&$55.80$&$-$&$14.65$&$34.90$&$45.55$&$56.95$\\
				SAN &$-$ & $14.35$ & $25.45$ & $43.35$ &$52.50$& $-$&$17.65$&$34.15$&$ 51.40$&$57.15$\\
				ETN &$-$ & $18.63$ & $27.76$ & $44.74$ &$58.13$& $-$&$15.76$&$32.30$&$ 48.30$&$56.61$\\
				\hline
				AIDA(ours) & \boldmath$91.30$ & \boldmath$92.65 $ & \boldmath$93.55$ & \boldmath$91.45$ & \boldmath$90.10 $ &  \boldmath$92.80 $ & \boldmath$93.65 $ &\boldmath$91.70 $ & \boldmath$92.55 $&  \boldmath$90.40$\\
				\hline
			\end{tabular}
	\end{minipage}}
	\par
	\subfigure[Macro-F1 comparison results]{
		\begin{minipage}{1.0\linewidth}
			\center
			\begin{tabular}{c|ccccc|ccccc}
				\hline
				Task &  \multicolumn{5}{c|}{$Article\quad Prediction$}   &\multicolumn{5}{c}{$Charge\quad Prediction$} \\
				\hline
				number & 50 & 100 & 200 & 500 & 1000 & 50 & 100 & 200 & 500 & 1000 \\
				\hline
				BiLSTM & $67.75$ & $74.01$ & $78.29$ & $80.77$ & $80.74$&$70.90$ & $74.78$ & $79.88$ & $81.57$&$81.09$\\
				TOPJUDGE & $78.55$ & $81.43$ & $82.11$ & $83.41$ & $83.33$&$80.29$ & $80.70$ & $81.83$ & $82.85$&$82.10$\\
				DDC & $67.72$ & $74.28$ & $76.27$ & $78.94$ & $77.40$ & $70.64$ & $74.75$ & $75.61$ & $80.09$ & $80.18$ \\
				DAN & $67.47$ & $74.76$ & $76.00$ & $83.97$ & $84.09$ & $71.20$ & $78.74$ & $81.69 $ & $83.29$ &$83.49$\\
				DANN& $70.28$ & $77.31$ & $74.79$ & $83.02$ & $83.09$ & $71.27$ & $77.08$ & $78.80$ & $81.48$&$82.95$\\
				JAN & $74.68$ & $78.50$ & $80.20$ & $83.55$ &$85.19$& $75.57$&$ 80.05$&$ 83.80$&$ 84.77$&$85.58$\\
				CDAN & $90.53$ & $91.39$ & $88.98$ & $89.92$ &$88.64$& $91.61$&$92.90$&$ 91.61$&$ 89.31$&$88.61$\\
				\hline
				PADA&$-$&$5.86$&$12.04$&$19.10$&$24.46$&$-$&$6.44$&$13.50$&$19.09$&$24.21$\\
				SAN &$-$ & $4.15$ & $10.93$ & $19.43$ &$23.85$& $-$&$7.30$&$12.93$&$ 19.48$&$24.12$\\
				ETN &$-$ & $6.29$ & $13.72$ & $20.41$ &$23.37$& $-$&$8.05$&$14.15$&$ 19.55$&$24.74$\\
				\hline
				AIDA(ours) & \boldmath$91.96$ & \boldmath$92.75 $ & \boldmath$93.58$ & \boldmath$92.36$ & \boldmath$90.89 $ &  \boldmath$93.71 $ & \boldmath$94.38$ &\boldmath$92.39$ & \boldmath$93.04 $&  \boldmath$91.57$\\
				\hline
			\end{tabular}
			
	\end{minipage}}
	\caption{AIDA vs the other algorithms with different imbalance ratio in law dataset. (\boldmath$"-"$ means the model cannot converge)}
	\label{Tab. 3}
\end{table*}
\begin{table*}
	\subfigure[Acc. comparison results]{
	\begin{minipage}{0.5\linewidth}	
		\center
		\begin{tabular}{c|cccc}
			\hline
			number & 50 & 100 & 200 & 500   \\
			\hline
			BiLSTM & $25.31$ & $29.08$ & $42.38$ & $47.00$ \\
			DDC & $37.62$ & $38.69$ & $40.38$ & $42.08$   \\
			DAN & $30.38$ & $35.08$ & $38.77$ & $48.76$  \\
			DANN& $36.00$ & $38.00$ & $42.46$ & $53.15$  \\
			JAN & $41.53$ & $39.00$ & $39.31$ & $56.00$ \\
			CDAN & $45.00$ & $38.92$ & $69.31$ & $71.31$ \\
			\hline
			AIDA(ours) & \boldmath$68.69$ & \boldmath$54.23 $ & \boldmath$72.30$ & \boldmath$73.23$  \\
			\hline
		\end{tabular}
	\end{minipage}}
	\hfill
\subfigure[Macro-F1 comparison results]{
\begin{minipage}{0.5\linewidth}
	\center
	\begin{tabular}{c|cccc}
		\hline
		number & 50 & 100 & 200 & 500 \\
		\hline
		BiLSTM & $36.11$ & $37.84$ & $41.95$ & $42.73$ \\
		DDC & $38.73$ & $39.91$ & $41.88$ & $43.00$  \\
		DAN & $40.70$ & $41.87$ & $43.29$ & $44.06 $ \\
		DANN& $41.26$ & $42.00$ & $42.86$ & $43.12$ \\
		JAN & $40.70$ & $39.46$ & $41.02$ & $44.85$\\
		CDAN & $40.91$ & $43.92$ & $46.63$ & $48.00$\\
		\hline
		AIDA(ours) & \boldmath$43.68$ & \boldmath$45.56 $ & \boldmath$49.57$ & \boldmath$49.80$ \\
		\hline
	\end{tabular}

\end{minipage}}
\caption{AIDA vs the state-of-the-art algorithms with different imbalance ratio in topic classification dataset.}
\label{Tab. 4}
\end{table*}

\section{Experiments and discussion}
In this section, we compare the proposed AIDA with the-state-of-the-art algorithms to demonstrate the effectiveness of our model.
\subsection{Data Preparation}
For \textit{PFD} dataset, we use Chinese AI and Law Challenge(CAIL)\cite{xiao2018cail2018}, a criminal case dataset for competitions released by the Superme People's Court of China. Although CAIL does not give the corresponding hierarchical structure, with the \textbf{nature hierarchy of chapters in the China Civil Law System}, 75 subclasses and 83 subclasses are assigned to the corresponding 8 parent classes for article and charge prediction task respectively.\par
Since there doesn't exist an off-the-shelf benchmark \textit{NPFD} dataset for our task, we construct our dataset from scratch. We crawled the data in the criminal field from legal Q\&A communities \footnote{https://www.66law.cn/}. The crawled data only have \textit{the charge label}, so we asking 5 law school master students to annotate the \textit{law article label}. We set these labelled dataset of \textit{charges} and \textit{law articles} as ground-truth for our \textit{NPFD} predict tasks. We list the detailed statistics of these datasets in Table \ref{Tab.1}.\par
To further verify the feasibility of our approach, we tested on public datasets: We use the Web of Science dataset \cite{Kowsari2018HDLTex} as the source domain data. This dataset contains 46,985 documents with 134 categories which include 7 parents categories. Target domain data from recommending topics of each question in Quora\footnote{https://github.com/nishnik/QuoRecommender}. We extracted the target subcategories with intersections in the WOS dataset and the Quora dataset. We list the detailed statistics of these datasets in Table \ref{Tab.2}.
\subsection{Experimental Settings}
We employ THULAC\cite{sun2016thulac} for word segmentation for these Chinese law datasets. Then, we adopt the Skip-Gram model \cite{mikolov2013distributed} to pre-train word embeddings with embedding size 300 and frequency threshold set to 25. The size of hidden state of BiLSTM is set to 128, so after text encoding, a vector representation size is 256. We only add a full connection layer with size 256 before the task classier layer. For the domain adaptator we stick to the three fully connected layers(x $\rightarrow$ 500 $\rightarrow$ 500 $\rightarrow$ 2) with the learning rate as 0.01. Meanwhile, we set the maximum sentence length to 300 words. In our training, the Stochastic gradient descent is used. \par

To evaluate the model's sensitivity to the imbalance of the training data, we set the shared-class in \textit{PFD} domain with different imbalance ratios (the average size of shared-classes in \textit{PFD} with randomized sampling), such as 50, 100, 200, 500, 1000.
\subsection{Evaluation metrics}
For the law classification and topic classification tasks, both of source domain are imbalanced. So We employ accuracy (Acc.) and macro-F1(F1) to evaluate the performance of the baselines and our methods, where the macro-F1 is the average F1 of eachcategory\cite{zhong2018legal}.
\subsection{Algorithms for Comparison}
We compare AIDA with two setting of state-of-the-art transfer methods such as shared domain adaptation setting and partial domain adaptation setting.\par
\textbf{Shared domain adaptation setting:} \par
(1)BiLSTM: The BiLSTM model is trained on the source domain with class labels and then transfer parameters of all layers from the source domain to the target domain.\par
(2)TOPJUDGE\cite{zhong2018legal}: Applying the
dependencies among the article and charge prediction as a form of DAG and introduce this prior knowledge to enhance judgment prediction. We also transfer parameters of all layers from the source domain to the target domain. \par
(3)DDC\cite{tzeng2014deep}:The DDC method used MMD in addition to the regular classification loss on the source to learn a representation that is both discriminative and domain invariant. \par
(4)DAN\cite{long2015learning}:The domain discrepancy is further reduced using an optimal multi-kernel selection method for mean embedding matching.\par
(5)DANN\cite{ganin2014unsupervised}:The gradient
reversal algorithm (ReverseGrad) proposed treats domain invariance as a binary classification problem, and directly maximizes the loss of the domain classifier by reversing its gradients.\par
(6)JAN\cite{long2017deep}:The proposed approach uses a JMMD penalty to reduces the shift in joint distributions of the network activations of multiple task-specific layers, which approximates the shift in the joint distributions of input features and output labels.\par
(7)CDAN\cite{long2018conditional}:The proposed CDAN method uses  multilinear conditioning that captures the cross-covariance between feature representations and classifier predictions to improve the discriminability and entroy conditioning that controls the uncertainty of classifier predictions to guarantee the transferability.\par
\textbf{Partial domain adaptation setting:} \par
(8)PADA\cite{Cao2018}:The proposed method employing only one adversarial network and further adding the class-level weight to the source classifier.\par
(9)SAN\cite{Cao_2018_CVPR}:This paper adopts multiple adversarial networks with weighting mechanism to select out source examples in the non-share classes.\par
(10)ETN\cite{Cao_2019_CVPR}:The author use a progressive weighting scheme to quantify the transferability of source examples.\par

\subsection{Results}
The law classification results on the Article Prediction and Charge Prediction are shown in Table \ref{Tab. 3}(a) and \ref{Tab. 3}(b), and the topic classification results are shown in Table \ref{Tab. 4}(a) and \ref{Tab. 4}(b). \textbf{The AIDA model for partial-and-imbalanced domain adaptation outperforms the previous state-of-the-art algorithms for shared domain adaptation and partial domain adaptation.} In particular, AIDA consistently improves the accuracy by \textbf{huge margins on tasks with large imbalanced ratio} in source domain, when the shared classes only have 50 samples per class. And it achieves considerable improvement on tasks when the source shared classes have 1000 samples in law classification task and the source shared classes have 500 samples in topic classification task. These results suggest that AIDA is promising to learn transferable features for the partial-and-imbalanced transfer learning in all tasks under the setting where target label space is a subset of the source lab space as well as the source domain is imbalanced.  \par
The results reveal some interesting observations. (1) The more data in source domain, the more classification accuracy for \textbf{shared domain adaptation algorithms}. (2) Partial domain adaptation algorithms with the assumption of \textbf{unknown target label space} are \textbf{worse} than shared domain adaptation, especially worse for imbalanced situation, large-scale non-shared situation or cannot convergence in 50 setting.  (3) Non-shared class knowledge, especially sibling classes of the shared ones, can be borrowed to improve the \textbf{partial-and-imbalanced domain adaptation} algorithm. The more imbalanced of the source dataset, the more relatively improvement can be achieved through borrowing sibling and non-shared classes knowledge via AIDA algorithm.

\subsection{Analysis}
\begin{figure}[t]
	\subfigure[ $\lambda_2=0.2$]{
		\begin{minipage}[t]{0.48\linewidth} 
			\centering
			\includegraphics[width=1.72in]{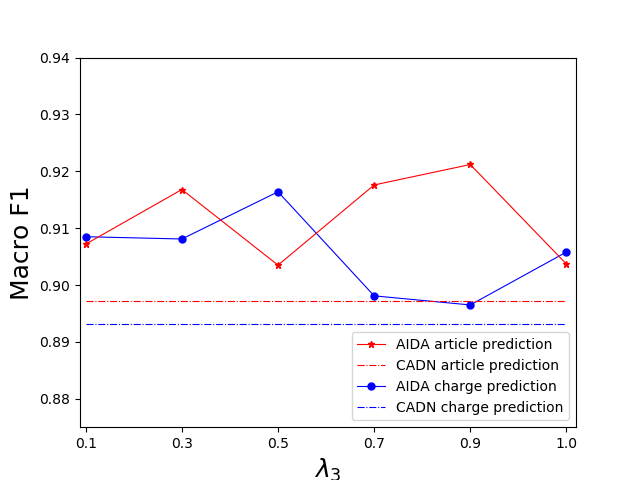}
	\end{minipage}}
	\hfill  
	\subfigure[ $\lambda_2=0.4$]{
		\begin{minipage}[t]{0.48\linewidth} 
			\centering
			\includegraphics[width=1.72in]{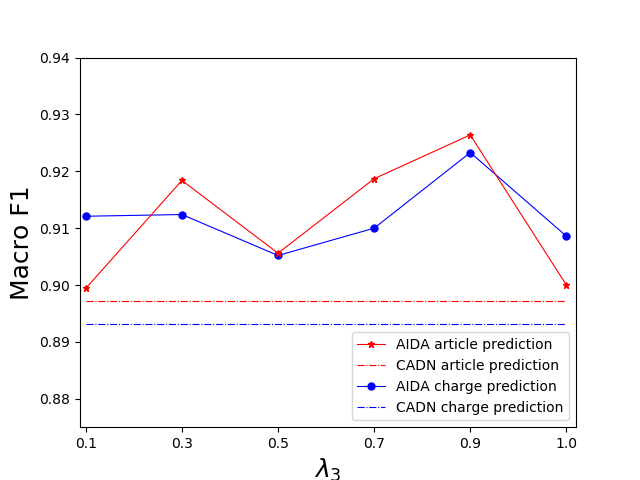}
	\end{minipage}}
\subfigure[ $\lambda_2=0.6$]{
	\begin{minipage}[t]{0.48\linewidth} 
		\centering
		\includegraphics[width=1.72in]{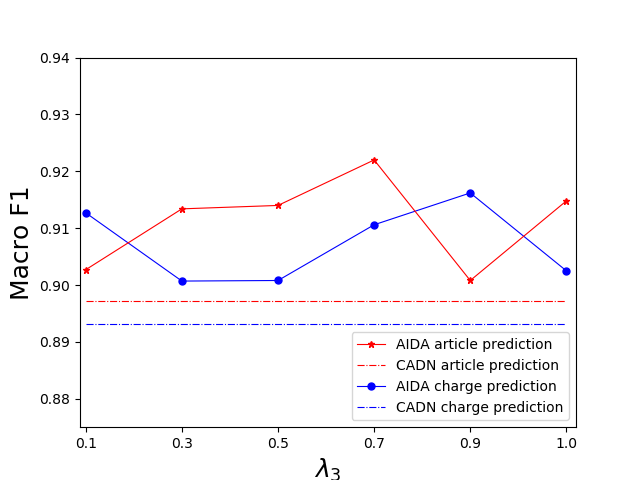}
\end{minipage}}
\subfigure[ $\lambda_2=0.8$]{
	\begin{minipage}[t]{0.48\linewidth} 
		\centering
		\includegraphics[width=1.72in]{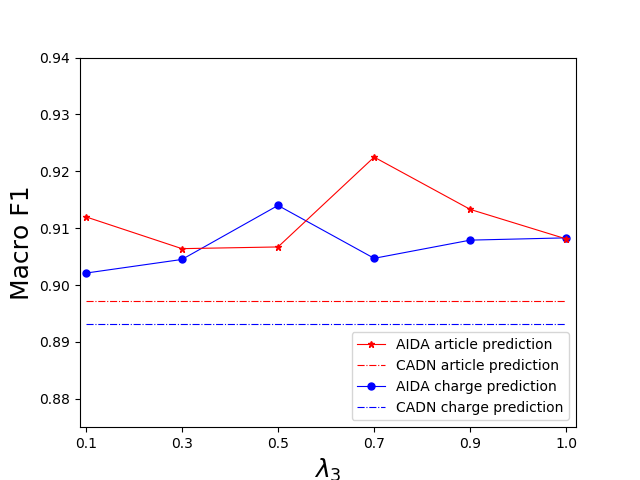}
\end{minipage}}
\subfigure[ $\lambda_2=1.0$]{
	\begin{minipage}[t]{0.48\linewidth} 
		\centering
		\includegraphics[width=1.72in]{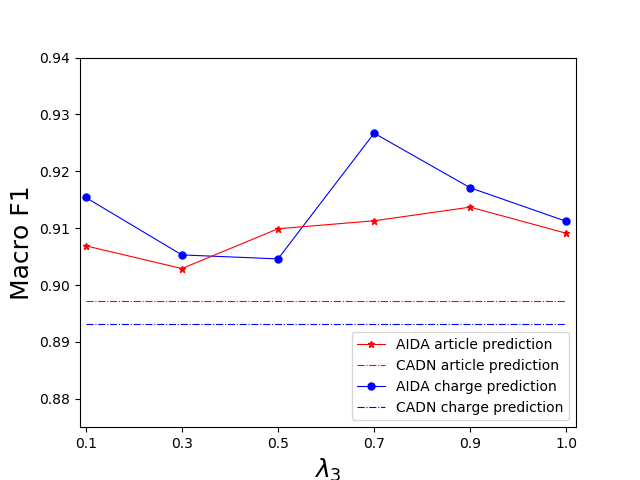}
\end{minipage}}
\subfigure[ $\lambda_2=2.0$]{
	\begin{minipage}[t]{0.48\linewidth} 
		\centering
		\includegraphics[width=1.72in]{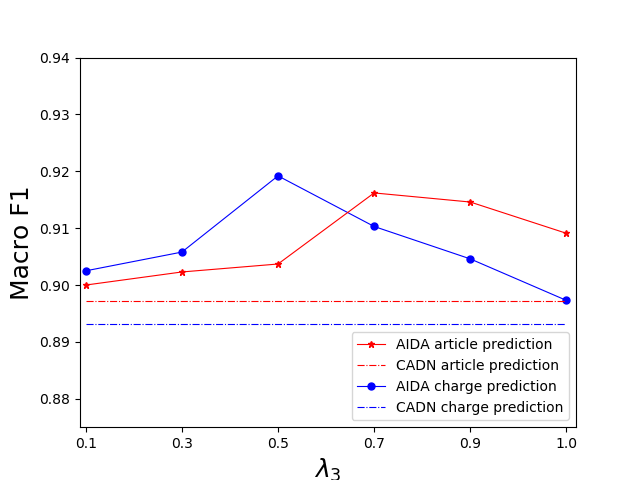}
\end{minipage}}
	\caption{Parameter Sensitivity} 
	\label{Fig. 4}
\end{figure}
\begin{figure}[t]
		\subfigure[Article prediction task]{
			\begin{minipage}[t]{0.48\linewidth}
				\centering
				\includegraphics[width=1.74in]{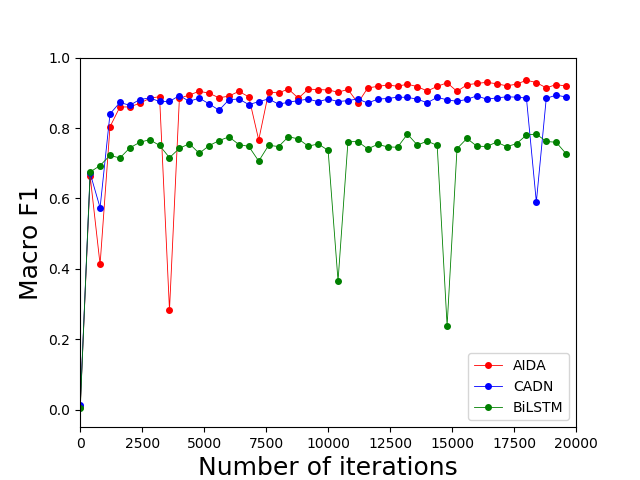}
		\end{minipage}}
	\subfigure[Charge prediction task]{
		\begin{minipage}[t]{0.48\linewidth}
			\centering
			\includegraphics[width=1.74in]{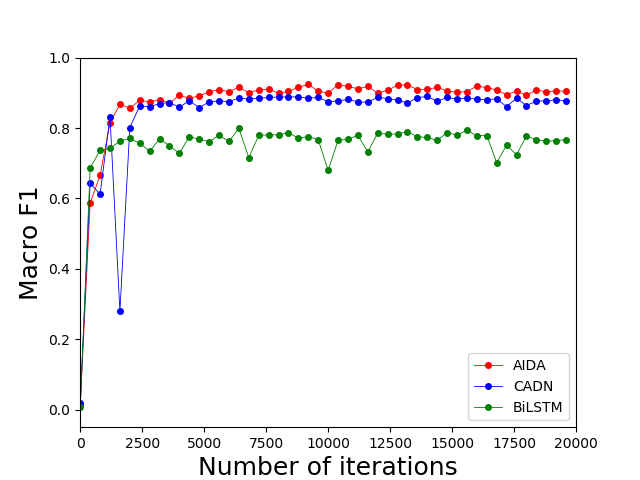}
	\end{minipage}}
	\caption{Convergence Performance} 
	\label{Fig. 5}
\end{figure}
\textbf{Parameter Sensitivity.}:
According to the parameters setting in the paper \cite{long2018conditional}, we set the $\lambda$ as 1. We investigate a broader scope of our algorithm AIDA for law article predictions and charge predictions by varying parameters  $\lambda_2$ from 0.2 to 2.0 and $\lambda_3$ from 0.1 to 1.0.
The results are shown in Fig \ref{Fig. 4}(a),\ref{Fig. 4}(b),\ref{Fig. 4}(c),\ref{Fig. 4}(d),\ref{Fig. 4}(e), and \ref{Fig. 4}(f). In these pictures, we fixed the values of $\lambda_2$ to [0.2,0.4,0.6,0.8,1.0,2.0] and then adjusted the value of $\lambda_3$=[0.1,0.3,0.5,0.7,0.9,1.0]. We have the following observations: (1) the best parameters of  $\lambda_2$ and $\lambda_3$ for charge prediction and article prediction tasks are all 0.4 and 0.9. (2) Even though we set different values for $\lambda_2$ and $\lambda_3$, all performance is higher than previous algorithms, which indicates that our proposed algorithm has a good stability.

\textbf{Convergence Performance}: We testify the convergence of AIDA and baseline algorithms of BiLSTM and CADN by studying the Macro-F1 through the training procedure.  Fig \ref{Fig. 5}(a) and Fig \ref{Fig. 5}(b) shows the results on article prediction and charge prediction tasks when the shared classes have 200 samples per class in \textit{PDF} domain. It indicates that \emph{AIDA} has a similar convergence speed as CADN. However, \textbf{the performance of AIDA is significantly higher than CADN}.
\par

\begin{figure}[t]
	\subfigure[Error Rate]{
		\begin{minipage}[b]{0.48\linewidth} 
			\centering
			\includegraphics[width=1.72in]{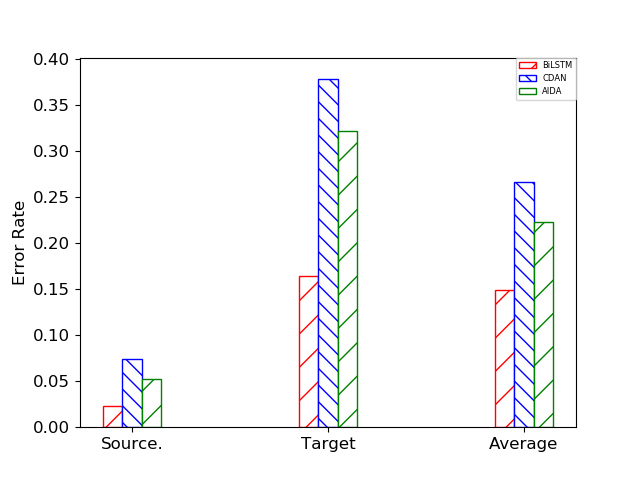}
	\end{minipage}}
	\hfill  
	\subfigure[A-distance]{
		\begin{minipage}[b]{0.48\linewidth}
			\centering
			\includegraphics[width=1.72in]{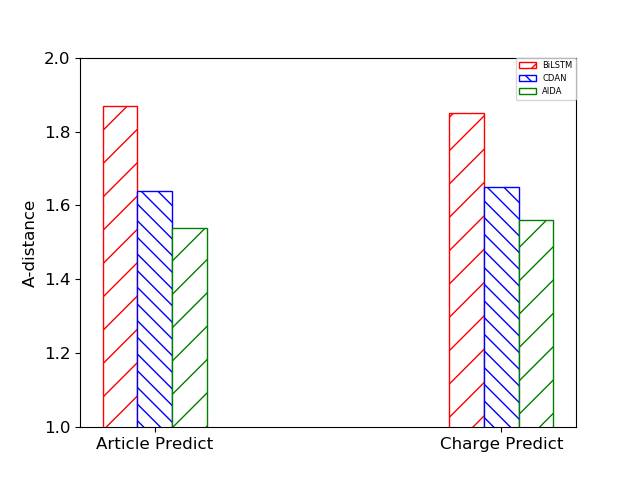}
	\end{minipage}}
	\caption{Adaptability and Discrepancy of learned features: (a) Ideal classification error rate; (b)A-distance} 
	\label{Fig. 6}
\end{figure}
\textbf{Adaptability Error.} We study the adaptability error, which can be built by training an MLP classifier on shared \textit{PFD} data and target \textit{NPFD} data with labels. The adversarial domain adaptation may reduce this ideal classification error since the over-distorting for learned features, which is defined as $\lambda^\star$ in Equation (27). The results are shown in Fig. 6(a). As a result, the adaptability $\lambda^\star$ of pre-trained ResNet is much lower than all domain adversarial networks, as well as \textbf{AIDA} significantly improve the adaptability of \textbf{CDAN} since weighted hierarchy adaptation.
\par
\textbf{Domain Discrepancy.} Domain discrepancy can be measured by the A-distance \cite{ben2010theory}, calculated with $2(1-2\epsilon)$, where $\epsilon$ is the classifier error trained for shared \textit{PFD} domain and target \textit{NPFD} domain. Results on the most sparse tasks of \textit{Article Prediction} and \textit{Charge Prediction} are shown in Fig. 6(b). The domain discrepancy of \textbf{AIDA} is smaller than \textbf{CDAN}. These two experiments show AIDA can improve the adaptability by reducing distorting as well as the domain discrepancy in sparse case by sibling constraint.
\par

\begin{figure}[htbp]
	\includegraphics[height=4.5cm,width=9.0cm]{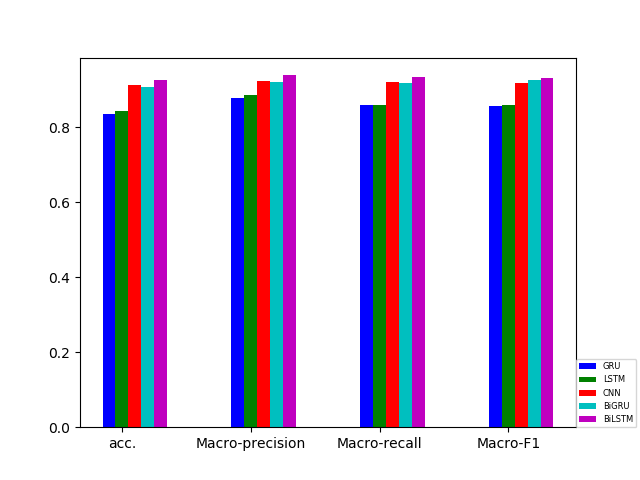}
	\caption{Compare the performance of different feature Generators}
	\label{Fig. 7}
\end{figure}
\textbf{Text Feature Generator.}
We further compare the performance of different feature extractors: CNN\cite{kim2014convolutional},
LSTM\cite{hochreiter1997long}, GRU\cite{cho2014learning}, BiGRU. Fig \ref{Fig. 7} shows the results on charge prediction task when the shared classes have 500 samples per class in \textit{PDF} domain.  We can draw the following points: (1) The performance of GRU is slightly lower than LSTM, and BiGRU/BiLSTM is higher than GRU/LSTM respectively. (2)The performance of CNN is higher than GRU/LSTM, but lower than BiGRU/BiLSTM. An important reason is that BiGRU/BiLSTM can better capture text features from left to right and enhance network migration capability.
\par
\textbf{Case Study.}: We select some cases to illustration the significance of AIDA:\par (1)"(What is the limit amount of cutting down fruit trees according to the amendment to the Criminal Law?)". The actual charge label is "(The crime of illegally logging)" and article label is legal provisions of law article 345. AIDA predicts all judgments correctly, while CADN fails to predict the charge label as "(The crime of intentional injury)" and the article label as legal provisions law article 133.\par(2)"(My uncle is an official. He received 200,000 yuan from the subordinate and promoted him recently. However, he was reported yesterday. I want to know what crime my uncle will be convicted of.)". The actual charge label is "(Bribery)" and article label is legal provisions of law article 385. AIDA predicts all judgments correctly, while CADN fails to predict the charge label as "(The crime of obstructing public servants)" and the article label as legal provisions law article 347.
\par

\section{Conclusion}
In this paper, we present a novel approach (\emph{AIDA} model) for the partial-and-imbalanced domain adaptation for \textit{NPFD}. Unlike previous GAN-based domain adaptation methods that exactly match the labels for both domains, the proposed approach simultaneously solves over-fitting of shared-classes by borrowing sibling knowledge in non-shared classes of \textit{PDF} domain and improve shared classes domain adaptation by hierarchy weighting adaptation network. Our approach successfully solve partial-and-imbalanced domain adaptation problem where source label space subsumes target label space and the source domain is imbalanced, evaluated by extensive experiments on real-world law datasets. This effective AIDA solution is a key step in practice for automatically predicting \textit{the charge} and \textit{the article} for NPFD in Q\&A community.

\section*{Acknowledgment}

The authors would like to thank...

\ifCLASSOPTIONcaptionsoff
  \newpage
\fi



\bibliographystyle{IEEEtran}
\bibliography{AIDA}
%
%
%

%

\begin{IEEEbiography}[{\includegraphics[width=1in,height=1.25in,clip,keepaspectratio]{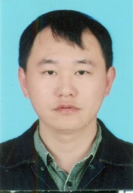}}]{Guangyi Xiao}
Hunan University, Changsha, China

Guangyi Xiao received the B.Econ. degree in software engineering from the Hunan University, Changsha, China, in 2006, the M.Sc. degree in software engineering from the University of Macao, Macao, China, in 2009 and the Ph.D. degree in software engineering from University of Macao, Macao, in 2015.

His principal research is in the field of partial-and-imbalanced transfer learning, computer vision, semantic representation, semantic integration, interoperation, and collaboration systems, mainly applied to the fields of food-computing, law-computing, e-commerce, e-marketplace.
\end{IEEEbiography}

\begin{IEEEbiography}[{\includegraphics[width=1in,height=1.25in,clip,keepaspectratio]{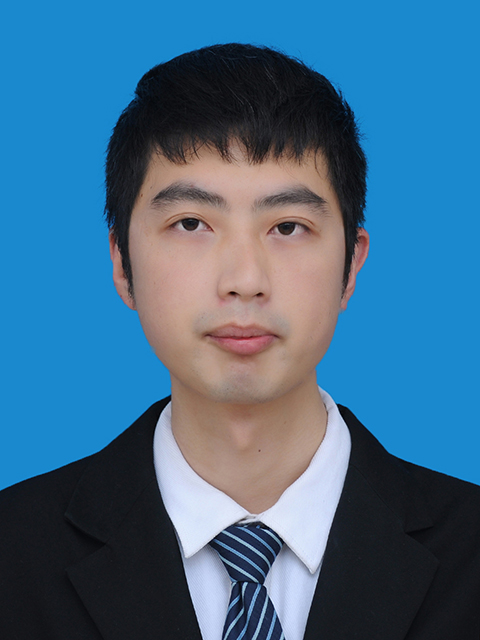}}]{Xinlong Liu}
Xinlong Liu received the B.Econ. degree in communication engineering from the Northwest University , Xian, China, in 2016.
\end{IEEEbiography}

\begin{IEEEbiography}[{\includegraphics[width=1in,height=1.25in,clip,keepaspectratio]{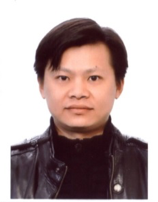}}]{Hao Chen}
Hunan University, Changsha, Hunan, China

Hao Chen received the M.S. degree in software engineering from Hunan University, Changsha, China, in 2004, and the Ph.D. degree from the School of Information Science and Engineering, Changsha, China, in 2012.

He is currently an Associate Professor and Ph.D. Supervisor at the College of Computer Science and Electronic Engineering, Hunan University, Changsha, China. His research interests are Web mining, personalized recommendation and big data technology.
\end{IEEEbiography}

\begin{IEEEbiography}[{\includegraphics[width=1in,height=1.25in,clip,keepaspectratio]{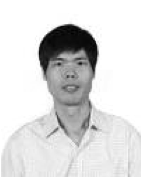}}]{Jingzhi Guo}
University of Macau, Taipa, Macau, China

Jingzhi Guo (M'05) received his PhD degree in Internet computing and e-commerce from Griffith University, Australia in 2005, the MSc degree in computation from the University of Manchester, UK, and the BEcon degree in international business management from the University of International Business and Economics, China in 1988.

He is currently an Associate Professor in e-commerce technology with University of Macau, Macao. His principal researches are in the fields of semantic integration, virtual world and e-commerce technology.
\end{IEEEbiography}

\begin{IEEEbiography}[{\includegraphics[width=1in,height=1.25in,clip,keepaspectratio]{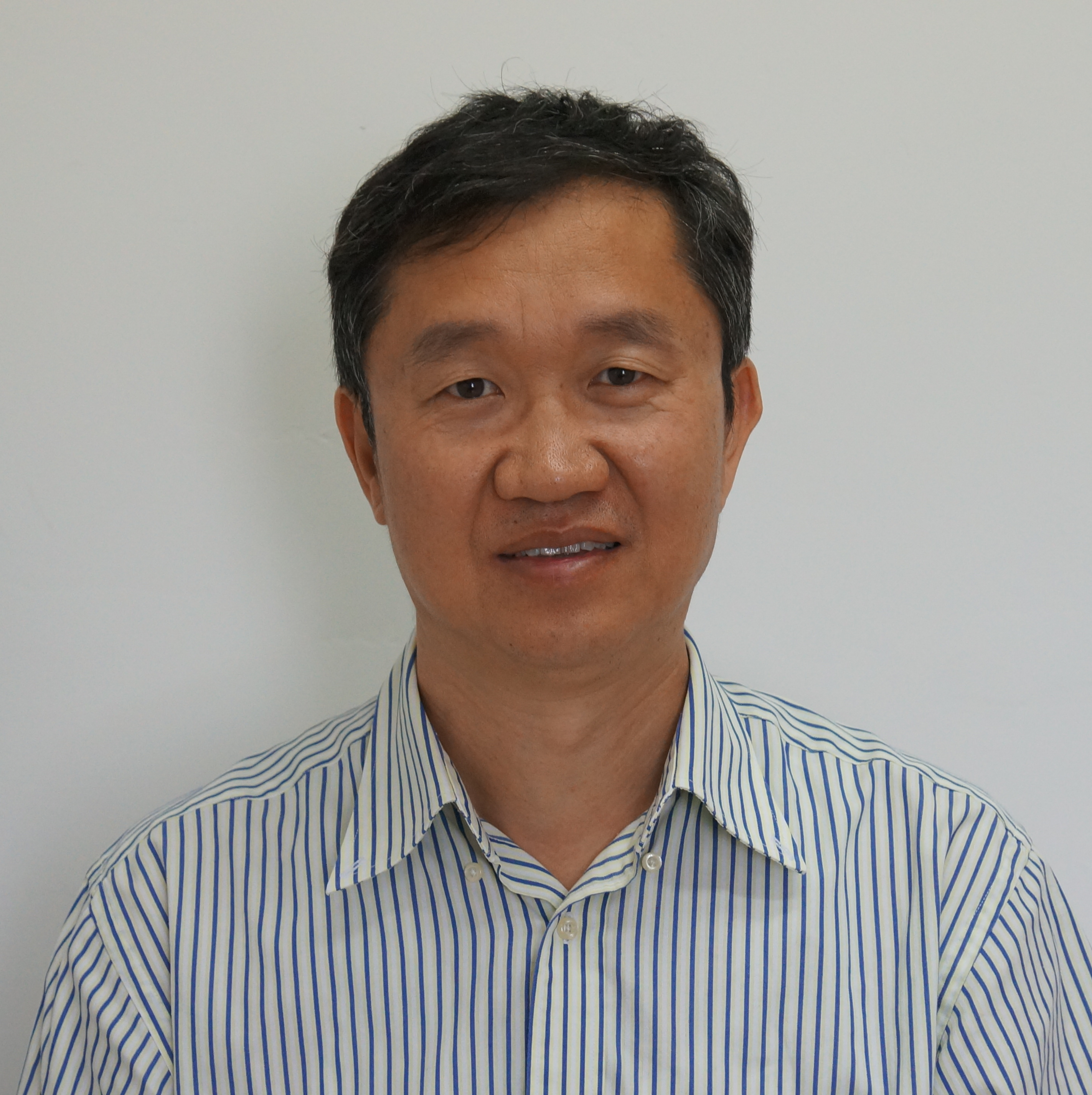}}]{Zhiguo Gong}
University of Macau, Taipa, Macau, China

Zhiguo Gong received the PhD degree in the Department of Computer Science, Institute of Mathematics, Chinese Academy of Science, and the MSc degree from Peking University, Beijing, China, in 1988.

He is currently an Professor and the Head in the Department of Computer and Information Science, University of Macau, Macau, China. His research interests include Machine Learning, Data Mining, Database, and Information Retrieval. He is a senior member of the IEEE.
\end{IEEEbiography}







\end{document}